\documentclass[sigconf,natbib=true,anonymous=false]{acmart}
\usepackage[inline]{enumitem}
\usepackage{caption}
\usepackage{subcaption}
\usepackage{multirow}
\usepackage{amsmath}
\usepackage{balance}

\AtBeginDocument{%
  \providecommand\BibTeX{{%
    \normalfont B\kern-0.5em{\scshape i\kern-0.25em b}\kern-0.8em\TeX}}}

\copyrightyear{2022}
\acmYear{2022}
\setcopyright{acmlicensed}
\acmConference[CIKM '22]{Proceedings of the 31st ACM International Conference on Information and Knowledge Management}{October 17--21, 2022}{Atlanta, GA, USA}
\acmBooktitle{Proceedings of the 31st ACM International Conference on Information and Knowledge Management (CIKM '22), October 17--21, 2022, Atlanta, GA, USA}
\acmPrice{15.00}
\acmDOI{10.1145/3511808.3557340}
\acmISBN{978-1-4503-9236-5/22/10}
\begin{document}

\title[GROWN+UP]{GROWN+UP: A “Graph Representation Of a Webpage” Network Utilizing Pre-training}

\author{Benedict Yeoh}
\affiliation{%
  \institution{Klass Engineering \& Solutions}
  \country{Singapore}
}
\email{benedict.yeoh@klasses.com.sg}

\author{Huijuan Wang}
\affiliation{%
  \institution{Klass Engineering \& Solutions}
  \country{Singapore}
}
\email{huijuan.wang@klasses.com.sg}

%%
%% By default, the full list of authors will be used in the page
%% headers. Often, this list is too long, and will overlap
%% other information printed in the page headers. This command allows
%% the author to define a more concise list
%% of authors' names for this purpose.
\renewcommand{\shortauthors}{Benedict Yeoh \& Huijuan Wang}

%%
%% The abstract is a short summary of the work to be presented in the
%% article.
\begin{abstract}
Large pre-trained neural networks are ubiquitous and critical to the success of many downstream tasks in natural language processing and computer vision. However, within the field of web information retrieval, there is a stark contrast in the lack of similarly flexible and powerful pre-trained models that can properly parse webpages. Consequently, we believe that common machine learning tasks like content extraction and information mining from webpages have low-hanging gains that yet remain untapped.

We aim to close the gap by introducing an agnostic deep graph neural network feature extractor that can ingest webpage structures, pre-train self-supervised on massive unlabeled data, and fine-tune to arbitrary tasks on webpages effectually.

Finally, we show that our pre-trained model achieves state-of-the-art results using multiple datasets on two very different benchmarks: webpage boilerplate removal and genre classification, thus lending support to its potential application in diverse downstream tasks.
\end{abstract}

%%
%% The code below is generated by the tool at http://dl.acm.org/ccs.cfm.
%% Please copy and paste the code instead of the example below.
%%
%\begin{comment}


%%
%% Keywords. The author(s) should pick words that accurately describe
%% the work being presented. Separate the keywords with commas.
%\end{comment}

\begin{teaserfigure}
    \centering
    \includegraphics[width=0.75\textwidth]{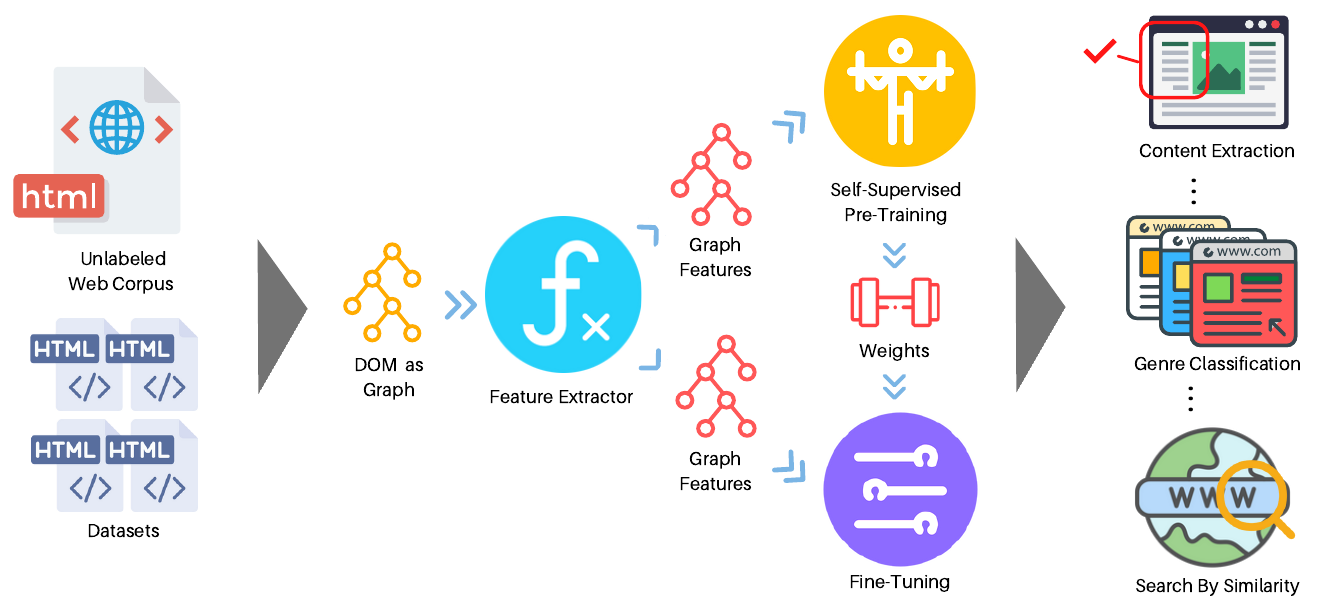}
    \caption{GROWN+UP at a glance}
    \Description{GROWN+UP illustrated in brief}
    \label{fig:teaser}
\end{teaserfigure}

\maketitle

\section{Introduction}
Deep convolutional networks have been a reliable cornerstone of computer vision (CV) in no small part due to the breakthrough results of AlexNet \cite{AlexNet2012} in 2012. Since then, a laundry list of inspired feature extractors in this space\cite{ResNet2015,MobileNet2017,VGG2015,InceptionNet2016} have gone on to apply transfer learning on image classification \cite{BiT2020,EfficientNetV22021}, object detection \cite{FasterRCNN2015}, instance segmentation \cite{ImageSegSurvey2021} and more \cite{TDN2021}, often with state-of-the-art results.

Similarly for natural language processing (NLP), recent innovations on pre-trained transformer-based feature extractors and their impressive results on several language modeling benchmarks \cite{BERT2019,GPT22019} have led to many other downstream applications \cite{NLPDLSurvey2019} and popular adoption.

%Indeed, the appeal of these pre-trained deep learning architectures is not solely attributed to their versatility across multiple downstream applications, but also partly because using pre-trained weights often lead to improved outcomes especially when training datasets are small \cite{ImageNetTransfer2019}\cite{Pretrain2020}. A pre-trained model can sometimes even perform competitively without any fine-tuning at all \cite{GPT22019}.

Then consider the domain of webpage information retrieval: By rough estimates, the number of webpages on the Internet is upwards of 4 billion in 2021 \cite{SearchEngineIndex2016,WWWSize2021}, representing an extremely rich repository of information for data mining. Correspondingly, there has been much published research into automated web content extraction, classification and related applications \cite{WebExtractSurvey2012}. These include vision-based approaches that use deep convolutional networks \cite{DNNWebExtract2016} to extract content.

%Contrasting webpage information retrieval with the CV and NLP domains, we note the rarity of a similar class of high capacity, pre-trained and conveniently versatile deep learning models that have seen much success in the latter fields

Yet, different from the CV and NLP domains, the class of high capacity, pre-trained and powerfully versatile feature extractors is still rare in the space of webpage information retrieval. This suggests that the beneficial trends seen in image and text parsing are not yet fully realized for webpage parsing, and by extension web information mining. We aim to exploit some of that potential by proposing GROWN+UP, a general model framework that comprises a webpage feature extractor based on a deep graph neural network (GNN) \cite{SpatialAndSpectral2020} that can be adapted for different tasks, coupled with a self-supervised approach to pre-train the model on unlabeled data.

In order to substantiate our claim that GROWN+UP can be flexibly applied, we choose two distinct tasks to benchmark on: web content extraction (specifically boilerplate removal) and genre classification. These two are selected because they exemplify tasks at the individual webpage element level and on the whole webpage respectively, each different enough to warrant specialized solutions. Moreover, being well-studied, there are sufficient prior works to compare against.

To the best of our knowledge, GROWN+UP is the first proposal for a general parser of webpages with supporting benchmarks on two traditionally distinct tasks. Specifically, our main contributions are summarized as follows:

\begin{itemize}
    \setlength\itemsep{0.5em}
    \item We introduce a general purpose GNN-based feature extractor that takes a webpage as input and extracts useful features for a variety of downstream tasks.

    \item We propose a self-supervised joint task prediction algorithm for pre-training on a large corpus of unlabeled webpages to learn generically applicable features.
    
    \item Due to incompatibilities and different standards used in past benchmarks on the boilerplate removal task, we re-evaluate past works on standardized dataset splits and use a direct similarity measure of the extracted text with the ground truth for a clear apples-to-apples comparison.

    %\item For the content extraction benchmark, we show that our pre-trained model is state-of-the-art. In particular on the Dragnet dataset \cite{DragnetData2021}, the improvement in the error from the previous best is around 50\%.

    %\item Finally, as support that GROWN+UP can be effective in various tasks, our general purpose model also posts competitive scores compared to other highly specialized classifiers in the genre classification benchmark.

    \item Finally, we verify that our general purpose model can attain top scores using reproducible measurements\footnote{\href{https://github.com/benyeoh/grownup}{https://github.com/benyeoh/grownup}} on two distinct and well-studied benchmarks, lending more credence to our claim.

    %\item Finally, as support that GROWN+UP can be highly effective in various tasks, we verify that our general purpose model can attain very respectable benchmark results on two dissimilar tasks using reproducible measurements. In particular, our model is state-of-the-art by a significant margin on the content extraction task.

    %\item For both the CleanEval \cite{CleanEval2008} and Dragnet \cite{DragnetData2021} datasets used in the content extraction benchmark, we show that our pre-trained model is state-of-the-art.

    %\item Finally, as validation that GROWN+UP can be applied effectively to different tasks, our general purpose model posts competitive scores compared to other highly specialized classifiers in the genre classification benchmark with the 7-Web-Genre \cite{Santini2008} and KI-04 \cite{KI_042004} datasets.
\end{itemize}

Each contribution is explained more thoroughly in the subsequent main sections.

\section{Methodology}
\subsection{Overview}
The GROWN+UP feature extractor takes a webpage HTML document and extracts feature representations that can be used for different tasks. %In addition to supporting predictions on the webpage, these features should also be granular enough to support predictions on individual webpage elements in order to account for tasks like content extraction.
Naturally, a few key questions need to be answered when devising this feature extractor, namely:

\begin{enumerate}
    \setlength\itemsep{0.5em}
    \item How do we practically represent a webpage for model ingestion?
    \item What is a suitable high-capacity model that can support tasks operating on individual webpage elements (like boilerplate removal) as well as the whole webpage?
    \item For this high-capacity model, how do we maximize performance and reduce over-fitting on small datasets?
\end{enumerate}

Our solution starts by considering the webpage DOM as a graph. This graph is then fed into a deep GNN-based model to extract features for every DOM element. Self-supervised pre-training is also employed to further boost performance generally and mitigate over-fitting.

In the following subsections, we elaborate on this approach and tackle those three questions in order.

\subsection{Model Inputs}

\begin{table}
    \caption{DOM Tag Input Features}
    \label{tab:input features}
    \small
    \begin{tabular}{lp{60mm}l}
        \toprule
        \normalfont{Name} & \normalfont{Description} & \normalfont{Size} \\
        \toprule
        text & USE embeddings of enclosed text \& length of text & 513\\
        class & USE embeddings of class attribute & 512\\
        id & USE embeddings of id attribute & 512\\
        tag type & one-hot vector of 83 most common HTML tags \& 1 catch-all & 84\\
        font weight & one-hot vector of HTML font weight defines: normal, bold and unknown & 3\\
        font style & one-hot vector of HTML font style defines: normal, italic, oblique and unknown & 4\\
        font size & one-hot vector of HTML font size defines: smaller, larger, xx-small, x-small, small, medium, large, x-large, xx-large and unknown & 10 \\
        %font size & one-hot vector of 10 classes: xx-small, x-small, small, medium, large, x-large, xx-large, smaller, larger and unknown & 10\\
        %font size & one-hot vector of 10 classes: smaller (<0.835em), larger (>1.2em), xx-small (1-9px), x-small (10-11px), small (12-14px), medium (15-17px), large (18-22px), x-large (23-30px), xx-large (>30px) and unknown & 10\\
        num child & number of children tags & 1 \\
        child index & one-hot vector representing child index from 0-30 \& 1 catch-all (>30) & 32\\
        pos encoding & graph Laplacian eigenvector values & 32\\
        \bottomrule
    \end{tabular}
\end{table}

%The GROWN+UP feature extractor takes a webpage HTML document as input and extracts a feature vector representation for downstream tasks.

Looking first at the webpage input, one of the key considerations is to preserve the original information content of the raw HTML during the pre-processing stage. That is, after encoding the HTML source into a more digestible form, we would like to properly represent not only the state of each HTML element in the document which can have different tag types and different tag attributes, but also the relationship between these elements. Preserving the original semantics instead of engineering features for a particular task will minimize inadvertent loss of useful information that could impact agnostic downstream decision making.

To start with, the webpage HTML DOM is viewed as a digraph (specifically a tree), where each HTML tag is a node in this graph and each directed edge is either of 3 types:
\begin{enumerate*}
    \item a node-to-parent edge, 
    \item a node-to-child edge or
    \item a node-to-self edge. 
\end{enumerate*} The direct parent-child relationship between HTML tags can thus be trivially represented. For purposes that will be clearer later, we also specify that every node in this graph has a node-to-self edge so that features from the source node are conveniently included in the neighborhood aggregation scheme of our model.

For practical reasons, our selection of graph node features are limited to a few DOM properties that we assume to be generically useful and do not require rendering nor additional remote resources to obtain. The selected features are detailed in Table \ref{tab:input features}.

%The selected features are
%\begin{enumerate*}
%    \item all text strings directly wrapped by the relevant tag,
%    \item the class attribute,
%    \item the id attribute,
%    \item the tag type,
%    \item the font size, weight and style,
%    \item the number of child tags,
%    \item the child index of the tag relative to parent, and lastly 
%    \item the positional encoding of the node from graph Laplacian eigenvectors % \cite{BenchmarkingGNN2020}\cite{LaplacianEigenmaps2003}.
%\end{enumerate*} More details are given in Table \ref{tab:input features}. 

\begin{figure*}
    \centering
    \includegraphics[width=\textwidth]{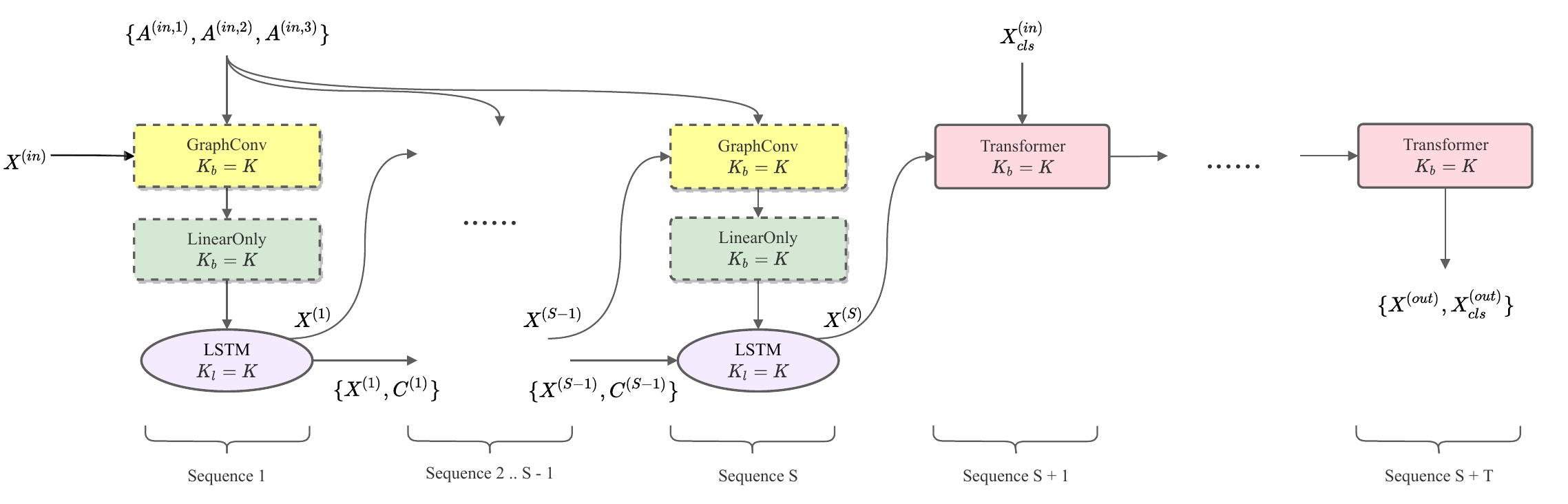}
    \caption{The feature extractor is composed of blocks (shown with dotted outlines) of component layers. $C^{(l)}$ are LSTM state matrices. The input graph edges are represented by 3 adjacency matrices, 1 per edge type.}
    \label{fig:GROWN+UP}
\end{figure*}

The text strings, class and id attributes are cross-lingual \cite{CrossLingual2019} sentence vectors generated using Universal Sentence Encoder (USE) \cite{USE2018, USEModel2021}. We choose to include the class and id attributes of a tag since it may contain useful semantic information \cite{Dragnet2013}, while the tag type and font properties were also shown to be useful for predicting relevant web content \cite{WebCE2013}.

% For the html tag type feature, we select 80 of the most common tags in our pre-training corpus for representation, with the other tags falling into a catch-all "unknown" class. Ideally, using the most common tags from the target corpus instead would likely give better results but we use the same set in all our experiments for convenience.

The number of child tags, child index and positional embeddings using graph Laplacian eigenvectors \cite{LaplacianEigenmaps2003,BenchmarkingGNN2020} give each node more spatial distinction relative to other nodes in the graph. Without positional information, neighborhood-aggregating GNNs typically cannot differentiate between subgraphs if the subgraphs exhibit structural symmetries \cite{HowPowerfulGNN2018}.

%Aside from the sentence vectors and positional embeddings of each node, all other properties such as the HTML tag type, font properties and the number of child tags are quantized and encoded as one-hot feature vectors.

%\setlength{\belowdisplayskip}{0pt} \setlength{\belowdisplayshortskip}{0pt}
%\setlength{\abovedisplayskip}{0pt} \setlength{\abovedisplayshortskip}{0pt}

Formally, we define graph inputs (and outputs) of some layer $l$ in our model to be composed of a graph node features matrix $X^{(l)} \in \mathbb{R}^{N \times K_{l}}$ and an adjacency matrix $A^{(l)} \in \{0, 1\}^{N \times N}$ where \[
    A_{ij}^{(l)} =
    \begin{cases}
        1, & \text{if directed edge } i \rightarrow j \text{ exists} \\
        0, & \text{otherwise} \\
    \end{cases}
\] and $N$ is the number of nodes of the graph and each $i$-th row $X_{i}^{(l)} \in \mathbb {R}^{K_{l}}$ of $X^{(l)}$ is a feature vector of size $K_{l}$ for node $i$. Slightly abusing notation, let $X^{(l)} + b$ also denote the matrix $(X_{1}^{(l)} + b, \; \cdots,\; X_{N}^{(l)} + b) \in \mathbb{R}^{N \times K_{l}}$ for some vector $b \in \mathbb {R}^{K_{l}}$. Subsequent descriptions of the model will refer to these definitions for conciseness.

\subsection{Model Architecture}

\begin{figure*}
    \centering
    \subcaptionbox{Transformer block.}% Aggregates 1-hop neighboring node features using the GraphConvGated layer.}%
        [.3\linewidth]{\includegraphics[height=0.35\textheight]{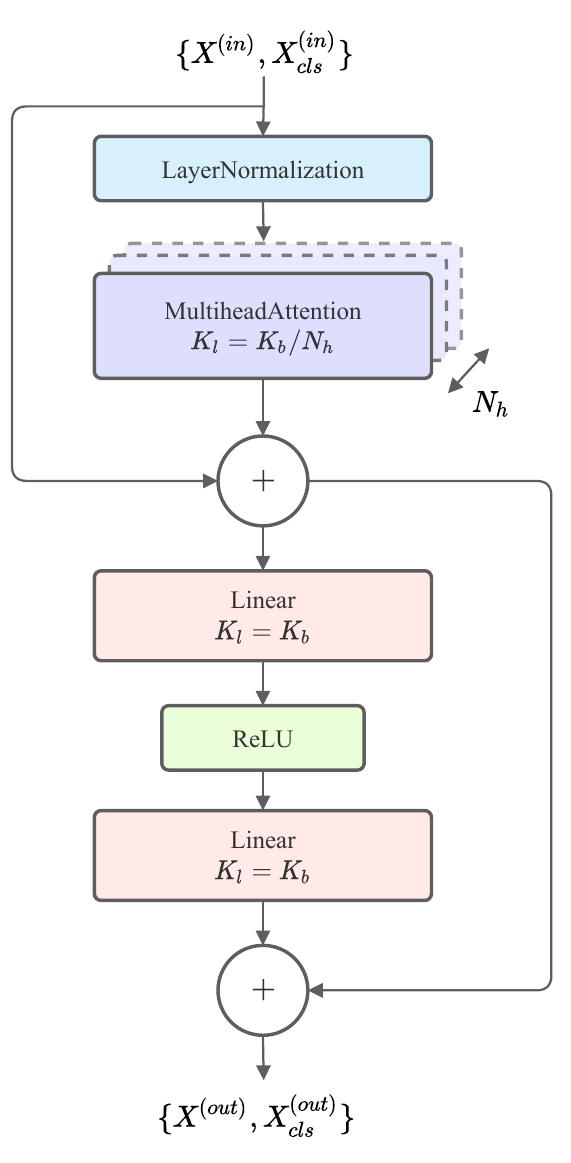}}\hfill
    \subcaptionbox{LinearOnly block.}% Similar in structure with GraphConv block but does not aggregate neighbours.}%
        [.3\linewidth]{\includegraphics[height=0.35\textheight]{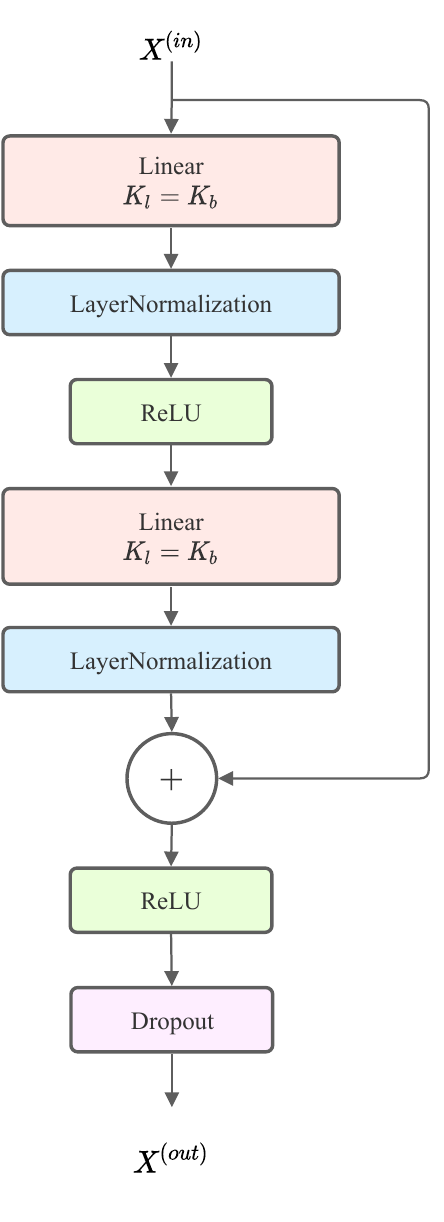}}\hfill
    \subcaptionbox{GraphConv block.}% Similar to LinearOnly block but also applies a projection on the residual. No dropout used.}
        [.3\linewidth]{\includegraphics[height=0.35\textheight]{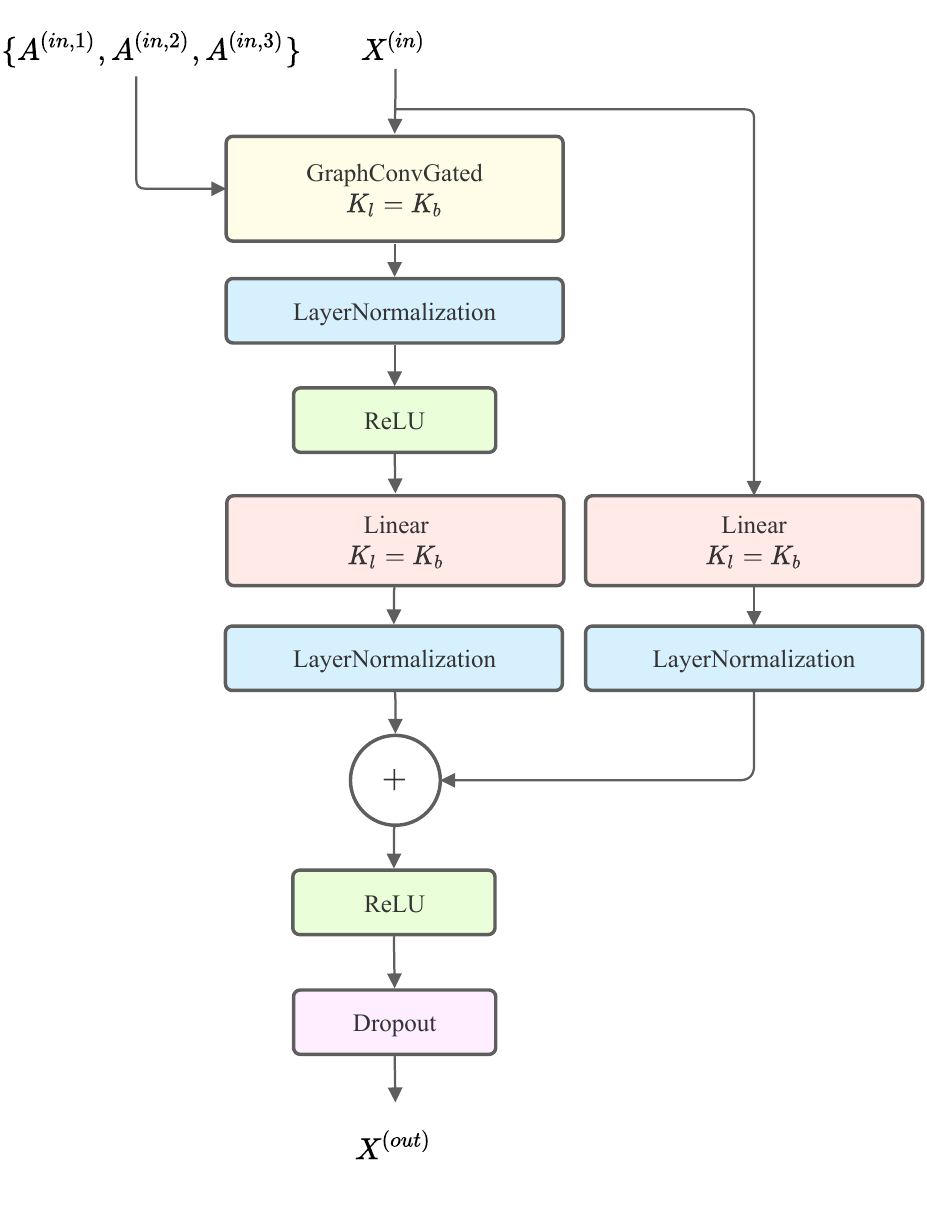}}
    \caption{Block component layers}
    \label{fig:Blocks}
\end{figure*}

In order to digest graph inputs and produce powerful feature representations per graph node, our feature extractor relies on stacking graph convolution operations in the style of Graph Convolution Networks (GCN) \cite{GCN2017}. Having features per node enables prediction tasks on the webpage element level, or on the whole webpage if aggregated, or in between.

Our feature extractor model also takes inspiration from deep CNNs such as ResNet \cite{ResNet2015} and bears some structural resemblance. Like ResNet, we assemble a deep network using a few blocks of repeating layers with skip connections to compute a residual mapping, albeit replacing 2D convolutions with 1-hop GCN-like graph convolutions or other operations. While it is known that piling on GCN layers exacerbates the over-smoothing problem \cite{GNNLoseExpressivePower2020} where the expressive power of the model is progressively lost, residual mappings have also been shown to mitigate this issue somewhat \cite{ResConvNet2017}.

One departure from a ResNet-esque design is the addition of an LSTM \cite{LSTM1997} layer after each graph convolution. Interestingly, this was also shown empirically to significantly improve the performance of very deep GCNs \cite{ResOrGate2019}. In the same study, the authors also showed that LSTM-augmented GCNs outperformed GCNs with residual mappings, but they did not compare a GCN using \emph{both} LSTMs and residual mappings. Particularly in our case, we found that using LSTMs and residuals together gives better empirical results over using one or the other exclusively. This observation suggests that the mechanisms by which both of these methods mitigate over-smoothing are somewhat complementary - at least in our architecture - and warrants further study.

Crucially, we also append Transformer \cite{Attention2017} blocks in the vein of Wu et al. \cite{Wu2021GraphTrans} to support feature aggregation on the whole graph (where edge information is ignored). This has been shown in their study to give superior results in graph-level classification tasks compared to simpler graph readout or pooling functions. Although we do not expect graph-level aggregation to benefit element-level tasks where node connectivity would likely play an important role, we still choose to use node features extracted from the output of Transformer blocks for element-level classification. We expect that unwanted noise can be mitigated \cite{ResNet2015} through the residual layers within each Transformer block.

On a high-level, the GROWN+UP feature extractor is organized as sequences of blocks, namely the
\begin{enumerate*}
    \item GraphConv,
    \item LinearOnly and
    \item Transformer
\end{enumerate*} blocks. These blocks are themselves abstractions of component layers which are basically self-contained operators such as graph convolutions or multi-head self-attention \cite{Attention2017}. As illustrated in Figure \ref{fig:GROWN+UP}, each sequence of GraphConv block, LinearOnly block and LSTM layer is stacked $S$ times to allow feature aggregation from neighbors $S$ hops away, with Transformer blocks appended $T$ times for long-range feature aggregation. We include dropout \cite{Dropout2014} after every GraphConv block and LinearOnly block. The feature size parameter $K$ controls the size of the output feature vectors of the block (denoted $K_{b}$). In practice, we default to $S = 5, T = 5, K = 256$ with number of attention heads $N_h = 4$. The various blocks and their component layers are shown graphically in Figure \ref{fig:Blocks}.

Lastly, besides the standard sigmoidal activations used within the LSTM layer, we maintain the use of ReLU \cite{ReLU2010} activations throughout.

In the following subsections, we elaborate on some of the core aforementioned layers for better clarity.

\subsubsection{GraphConvGated}
The GraphConvGated layer aggregates and processes features from neighboring nodes, roughly analogous to a 2D convolution in the case of images. We first give a brief overview of past works and then introduce our adaptation.

Kipf and Welling \cite{GCN2017} based their initial approximation of spectral graph convolutions on a 1st order Chebyshev polynomial expansion of the graph filter which they re-adjusted with a normalization scheme for better numerical stability. Finally, they ended up with a 1-hop neighborhood simplification that can be formulated as:

\begin{align*}
    X^{(l)}_i = \sum_{j \in \mathcal{N}(i)} \dfrac{1}{\sqrt{|\mathcal{N}(i)||\mathcal{N}(j)|}}X^{(in)}_j W
\end{align*} where $\mathcal{N}(u) = \{ v \: | \: A_{uv}^{(in)} = 1 \}$ are the neighboring nodes of source node $u$ and $W \in \mathbb{R}^{K_{in} \times K_{l}}$ are the learned parameters.

A side effect of this formulation is that neighborhood feature aggregation is isotropic - aside from a scalar normalization term, each neighbor is treated identically, potentially limiting the expressiveness of the model. This issue is made more apparent if we consider a regular grid and compare a standard 3x3 convolution kernel commonly used in image processing against the GCN formulation. Both of these operations have the same receptive field but the 3x3 kernel supports different sets of weights per neighboring node and is therefore much more expressive in that context. Nevertheless, the spectral basis behind the formulation of GCN remains theoretically appealing and its implementation is also straightforward.

To mitigate some of the aforementioned limitations, we leverage the aggregation scheme in Gated Graph ConvNet \cite{ResConvNet2017} which allows some degree of anisotropy through an edge gating mechanism for each source-neighbor pair. The claim is that this added gating with more learnable parameters should result in a more expressive model, and this seems consistent with some benchmarks where it is shown that gated aggregation outperforms other isotropic aggregation schemes \cite{BenchmarkingGNN2020}. We also adapt it to support a number of edge types where each edge type $k$ would have a distinct set of learned parameters to allow even greater flexibility.

More precisely, the output features of a node and the edge gates of the GraphConvGated layer are defined as:
\begin{gather*}
    X^{(l)}_i = \sum_{k} \sum_{j \in \mathcal{N}_{k}(i)} \eta_{ijk} \odot (X^{(in)}_j W_k + {b_k}), \\
    \eta_{ijk} = \sigma(X^{(in)}_{i}B_{k} + X^{(in)}_{j}C_{k})
\end{gather*} respectively, where $\sigma$ is the sigmoid function, $b_k \in \mathbb{R}^{K_{l}}$ is a learned bias vector, $W_{k}, B_{k}, C_{k} \in \mathbb{R}^{K_{in} \times K_{l}}$ are the learned gating weights and $\mathcal{N}_{k}(u) = \{ v \: | \: A_{uv}^{(in, k)} = 1 \}$ are the neighboring nodes of source node $u$ and edge type $k$.

\subsubsection{Linear}
While the GraphConvGated layer aggregates neighboring node information, the Linear layers only operate on individual node features with a simple linear transform. Specifically, the output node features matrix is defined by: 
\begin{gather*}
    X^{(l)} = X^{(in)}W + {b}
\end{gather*} where $W \in \mathbb{R}^{K_{in} \times K_{l}}$ and $b \in \mathbb{R}^{K_{l}}$ are the learned weights and bias respectively.

\subsubsection{MultiheadAttention}
Finally, within a Transformer block we use a configuration similar to Radford et al. \cite{GPT22019} with the output of the multi-head self-attention module defined as:
\begin{gather*}
    X^{(l)} = \text{concat}(H_1, \; H_2, \cdots, H_{N_h}) \\
\end{gather*} and 
\begin{gather*}
    H_{i} = \text{softmax}(\frac{H^{(in)}W_{q_i} (H^{(in)}W_{k_i})^{T}}{\sqrt{K_{in}}}) \cdot H^{(in)}W_{v_i}, \\
    H^{(in)} = (X_{1}^{(in)}, \; X_{2}^{(in)}, \cdots, X_{N}^{(in)}, X_{cls}^{(in)}) \\
\end{gather*}
where $N_h$ is the number of attention heads, $W_{q_i}, W_{k_i}, W_{v_i} \in \mathbb{R}^{K_{in} \times \frac{K_{l}}{N_h}}$ are self-attention weight tensors for head $H_i$. We also include a trainable feature embedding $X_{cls}^{(in)}$ in the input feature tensor that will be used for graph-level predictions \cite{Wu2021GraphTrans}.

\subsection{Self-Supervised Pre-training}
As the parameters and model capacity increases when we stack layers to make deeper networks, we pre-train on massive datasets to maximize generalization performance on downstream tasks and reduce the risk of overfitting on small datasets. Clearly, a self-supervised or unsupervised approach for pre-training would be advantageous since a purely supervised or even semi-supervised alternative would be extremely expensive to manually annotate at scale.

In formulating our approach, we take inspiration from a variety of sources including denoising autoencoders \cite{LayerWiseTraining2006,DenoiseAuto2010} and deep learning approaches to language modeling. In particular, the relation between sentence completion and general classification tasks formalized by Saunshi et al. \cite{MathLM2021} and the top performing BERT model \cite{BERT2019} are two works that were especially influential. %Very briefly, Saunshi et al. reformulated sentence completion in autoregressive language models into a general classification problem to explain some of the success of pre-trained language models in downstream applications. We refer the reader to their insightful work for more details.

We begin by considering BERT and noting their strategy to predict masked inputs has at least superficial resemblance to representation learning through a denoising criterion similar to denoising autoencoders. We may immediately adapt masked input prediction to our particular case, but a stronger intuition to justify that approach is desirable.

We proceed to offer an informal viewpoint relating BERT’s pre-training to downstream classification tasks with Saunshi et al.'s sentence completion reformulation as a backdrop, and build on that view to support our pre-training strategy.

BERT pre-trains on two objectives: \begin{enumerate*} \item cloze-like masked word prediction and \item next sentence prediction (NSP). \end{enumerate*} If we consider sentiment classification reformulated into a masked word prediction task, for example predicting “love” or “hate” in “I [MASK] exercise”, then pre-training on cloze-like masked word sentences can implicitly provide supervision for sentiment classification tasks. Furthermore, pre-training on the NSP objective seems on the surface relatable to the same sentence completion reformulation given by Saunshi et al. if the distribution of the predicted next word defined fundamentally in their work is replaced with the predicted next sentence instead. From that angle, the NSP objective seems to facilitate downstream classification likewise.

Now armed with a rough intuition for BERT's pre-training effect downstream, we broaden the notion of a sentence context in BERT to a more general graph context. Instead of predicting masked words, we predict masked DOM tag features. This is our first pre-training objective. The purpose is similar to our earlier reasoning for masked word prediction but adapted for DOM structures: we may contrive a case where such pre-training can implicitly assist some downstream \emph{element-level} classification tasks, such as sentiment analysis on social media posts, by constructing an appropriate DOM subtree resembling an addendum and predicting masked features for comparison against positive or negative templates.

Our second pre-training objective is a substitute for the NSP objective of BERT. We determine whether a pair of webpages are similar according to some chosen metric. The reason is as follows: Assuming that the model is able to determine if any two webpages are similar, then hypothetically we can classify a webpage by comparing it to several pre-defined template webpages from a downstream target class and choosing the closest match. Training on this objective should thus indirectly assist in some downstream \emph{webpage-level} classification tasks.

Procedural details for these element-level and webpage-level pre-training objectives are described next.

\subsubsection{Masked DOM tag feature prediction}
A webpage is represented as a graph where each node in the graph corresponds to a DOM tag element. For each input graph, we randomly select $M$ nodes for input feature prediction. In practice, we used $M = 16$. We then apply random masking on the selected nodes by zeroing all input features of those nodes with 85\% probability. However, since the input features of downstream applications are never completely zeros, masking introduces a disparity in the input distribution between pre-training and downstream tasks. This is mitigated by leaving a percentage unmasked, similar to the approach used in BERT. We predict multiple input features of the selected nodes simultaneously and optimize on the appropriate loss functions. The input features to predict are \begin{enumerate*} \item the encoded text wrapped by the tag, \item the tag type, \item the class attribute encoding, \item the id attribute encoding and lastly \item the number of child tags. \end{enumerate*}

For the tag type and number of child tags, we minimize the cross entropy loss given by:
\begin{gather*}
    L_{CE}(x, y) = - \sum_{i = 1}^{C} y_{i} \log{(x_{i})}
\end{gather*} where $x, y$ are the predicted and ground truth distributions respectively for some number of classes $C$. It is worth noting that in addition to the tag type, the number of child tags are encoded as one-hot vectors as well.

The feature embeddings for the wrapped text, class attribute and id attribute uses a cosine loss of the form:
\begin{gather*}  
    L_{cos}(x, y) = 1 - \dfrac{x \cdot y^{T}}{\|x\|\|y\|}
\end{gather*}

Let $\eta = \{\eta_{1}, \, \eta_{2},\, \cdots, \, \eta_{M} \: | \: \eta_{k} \in [1, N]\}$ be the indices of the $M$ randomly sampled and masked nodes and $X^{(l)}, Y^{(t)}$ be the feature extractor output and ground truth features respectively for some feature type $t \in T \coloneqq \{tag,\, nchild,\, text, \, class,\, id\}$. Then the loss $L_{t}(X^{(l)}, Y)$ per graph input is specified by:

\begin{gather*}
    L_{t_1}(X^{(l)}, Y) = \sum_{i=1}^{M} L_{CE}(X^{(t_1)}_{\eta_{i}}, Y^{(t_1)}_{\eta_{i}}), \\
              X^{(t_1)} = X^{(l)} W_{t_1} + {b_{t_1}}, \qquad \forall \, t _1\in \{tag, \, nchild\}
\end{gather*} and
\begin{gather*}
    L_{t_2}(X^{(l)}, Y) = \sum_{i=1}^{M} L_{cos}(X^{(t_2)}_{\eta_{i}}, Y^{(t_2)}_{\eta_{i}}), \\
              X^{(t_2)} = \text{tanh}(X^{(l)} W_{t_2} + {b_{t_2}}), \qquad \forall \, t_2 \in \{text, \, class, \, id\}
\end{gather*} where $W_{t \in T} \in \mathbb{R}^{K_{l} \times K_{t}}, \, b_{t \in T} \in \mathbb{R}^{K_{t}}$ are learned reprojection weights and biases.

\subsubsection{Same website prediction}
Typically, webpages that belong to the same genre should share some structural similarities in the DOM. For example, a product listing webpage is usually more similar in the visual layout and content to another product listing compared to a news article or a blog. Now suppose that we have a suitably varied dataset and a training objective that classifies webpages to its genre, then a trained model should also produce embeddings that are "close" for webpages in the same genre. In other words, a trained model is able to parse webpage similarities to some degree. We claim that these trained model features can also assist in other downstream webpage classification tasks. As mentioned earlier, we can make predictions on previously unseen classes by comparing webpages to predefined templates representing the target classes. An analogy can also be seen with pre-training on ImageNet classes and how it demonstrably benefits downstream classification in CV \cite{ImageNetTransfer2019,BiT2020}.

In practice however, getting labeled datasets suitable for training a webpage classifier for the purpose described is extremely expensive at scale. As an alternative, we instead compare a pair of webpages and predict if they originate from the same website, implicitly assuming that webpages from the same site share some structural similarities. This is a less-than-ideal approximation because it is entirely possible for webpages to be almost identical but hosted on different sites. Yet, it has the important advantage of obviating the need for manual labeling and requires no extra effort apart from grouping webpages by their website.

During pre-training, we first organize the inputs into pairs of webpages from the same website. Then we permute pairs of webpages in each batch such that roughly only half of the pairs are similar, generate appropriate binary labels, and finally optimize on a binary cross entropy loss. This algorithm has commonalities with negative sampling used in Word2Vec \cite{Word2Vec2013} where the negative samples in our case are drawn from a “noise” distribution created by permuting webpage pairs. Webpage feature vectors are derived from the CLS output of the last Transformer block (or alternatively from a graph readout function), followed by a hyperbolic tangent activation. More precisely, given the webpage feature vector outputs for a webpage pair, $X^{(l_1)}, \, X^{(l_2)}$ and input label $y \in \{0, 1\}$, the loss function $L_{sim}$ is defined as:

\begin{align*}%
    L_{sim}(X^{(l_1)}, X^{(l_2)}, y) = -y\log{z} - (1 - y)\log{(1 - z)} \,
\end{align*} with
\begin{gather*}
    z = \text{max} \{ \dfrac{\hat{x}_{1} \cdot \hat{x}_{2}^T}{\| \hat{x}_{1} \| \,  \| \hat{x}_{2} \|}, \, 0 \} \\
    \hat{x}_1 = \text{tanh}(\dfrac{1}{N} \sum_{i = 1}^{N} X^{(l_1)}_{i} W + b) \,,\quad \hat{x}_2 = \text{tanh}(\dfrac{1}{N} \sum_{i = 1}^{N} X^{(l_2)}_{i} W + b) \\
\end{gather*} where $W \in \mathbb{R}^{K_{l} \times K_{sim}}, \, b \in \mathbb{R}^{K_{sim}}$ are learned reprojection weights and biases.

Finally, the total weighted loss minimized during pre-training with experimentally-derived coefficients is given as:	
\begin{align*}
L & = 0.05 \: * \: L_{sim}(X^{(l_1)}, \, X^{(l_2)}, \, y) \\
& + 0.2 \: * \: [L_{tag}(X^{(l_1)}, Y^{(1)}) + L_{tag}(X^{(l_2)}, Y^{(2)})] \\
& + 0.5  \: * \: [L_{text}(X^{(l_1)}, Y^{(1)}) + L_{text}(X^{(l_2)}, Y^{(2)})] \\
& + 0.05 \: * \: [L_{id}((X^{(l_1)}, Y^{(1)}) + L_{id}(X^{(l_2)}, Y^{(2)})] \\
& + 0.1 \: * \: [L_{class}(X^{(l_1)}, Y^{(1)}) + L_{class}(X^{(l_2)}, Y^{(2)})] \\
& + 0.1 \: * \: [L_{nchild}(X^{(l_1)}, Y^{(1)}) + L_{nchild}(X^{(l_2)}, Y^{(2)})]
\end{align*}

These coefficients in general places more weight on predicting masked text embeddings and less on the same-website similarity measure, which in our experiments gave good results. In particular, we found that over-emphasizing the latter tends to result in slightly worse performance on our node level benchmarks (i.e. boilerplate removal).

\section{Related Work}

On the subject of pre-training, Hu et al. \cite{StratPreTrain2020} introduced node feature masking on GNNs in the style of BERT where they predict the masked node features from the surrounding context. This is essentially similar to one of our pre-training objectives. Our intuition leading to this approach however, is explained at length and inspired from Saunshi et al.’s \cite{MathLM2021}  sentence completion reformulation theories. Hu et al. also suggested a second stage of supervised pre-training for graph-level features after pre-training node-level features in the first stage. Our version of graph-level pre-training on webpages (i.e. same-website prediction) by contrast does not require manual labeling and can therefore scale to a massive amount of webpages inexpensively. Furthermore, unlike Hu et al.’s approach, we train both pre-training objectives simultaneously instead of sequentially in stages.

Next, given our selected benchmarks in boilerplate removal and genre classification, we briefly frame past works in these areas in relation to our proposed model.

In webpage genre classification, recent methods in literature run the gamut from traditional machine learning classifiers \cite{SVMWebClass2017,EnsembleWebClass2021,Santini2008} on hand-crafted webpage features to specialized deep learning architectures \cite{CompositeWebClass2019,ConvWebClass2021} and pre-trained language models \cite{BERTWebClass2020} that operate mostly on textual features within a webpage, but occasionally also on other sources such as the URL \cite{URLWebClass2005} and proprietary website descriptions \cite{WebClassDMOZ2015}. In this work, we focus only on webpage inputs, which naturally includes all textual features in addition to the rest of the DOM.

The task of extracting main textual content from webpages likewise also has a relatively long history, starting with heuristic based approaches such as Arc90's Readability application \cite{Readability2010}, Boilerpipe \cite{Boilerpipe2010}, CETD \cite{CETD} and CETR \cite{CETR} which use shallow text and hand engineered HTML tag features. Later, machine learning approaches such as Dragnet \cite{Dragnet2013} aggregated a variety of features, including features from past works such as CETR, and trained a classifier around those amalgam of features. More recent works like Web2Text \cite{Web2Text2018} and BoilerNet \cite{BoilerNet2020} employ deep learning to classify web content, but they still use an approach and feature set that remain coupled specifically to the web boilerplate removal task.

Broadly, most works on boilerplate removal operate from per-element features, such as features attached to each text block in the DOM \cite{Dragnet2013,Web2Text2018} or lines of text in the HTML \cite{CETR}. On the other hand, prior genre classification works normally consider webpage-level features such as term frequencies of words \cite{Santini2008}, n-grams \cite{Kanaris2009}, or pooled word embeddings \cite{ConvWebClass2021} of the entire document. In their respective prior works, there is little overlap between the tasks.

Rather than focus on boilerplate removal or webpage classification, GROWN+UP is distinct from prior works and can be applied to either of these tasks and more, being in the vein of general feature extractors seen in the CV and NLP domains.

Recently, MarkupLM \cite{MarkupLM2021} demonstrated that a BERT-based language model augmented with DOM XPath information and appropriate pre-training can attain state-of-the-art test scores on two webpage parsing datasets \cite{hao2011from, chen-etal-2021-websrc}. This is similar to GROWN+UP in the sense that MarkupLM is also an agnostic parser of webpages and can be applied to multiple tasks. However, being in essence a token-based \emph{language model} (albeit augmented with XPath embeddings per token) that excels in token-level text parsing and tasks, MarkupLM takes a very different approach from GROWN+UP towards webpage parsing. GROWN+UP is a webpage DOM model rather than a webpage-capable language model, and does not have fundamental dependencies on webpage textual content or lack thereof. Unlike MarkupLM, GROWN+UP can not only operate on webpages with no text but can also naturally handle webpage image or video input features, if included.

%To the best of our knowledge, GROWN+UP is the first proposal for a general parser of webpages with supporting benchmarks on two traditionally distinct tasks.

%By parsing DOM structures along with an appropriate choice of a cross-lingual sentence encoder \cite{USEModel2021} for textual features, our model can also be applied to a larger variety of webpages in different languages and even a mixture of languages. This is a huge benefit compared to prior works which mostly rely on hand-crafted English-only textual features \cite{BoilerNet2020}\cite{Web2Text2018}\cite{Santini2008}.

\section{Experiments}

\subsection{Self-Supervised Pre-training}
Pre-training requires a sufficiently large distribution of webpages from various sources in order to ensure good transferability to different domains. We randomly selected 180K webpages from the CommonCrawl\footnote{\url{https://commoncrawl.org/the-data/get-started/}} 2008 archive, and paired them based on matching URL subpaths for website similarity prediction. Here, we are making the reasonable assumption that webpages with similar URL subpaths belong to the same website.

Since webpages evolve over time, pre-training with webpages from a mix of time periods or ideally from the same time period as the target domain would likely garner the best results. %Also, we expect that using a more diverse corpus that is a couple orders of magnitude larger will also give greater improvements.
However, pre-training with our rather modestly sized CommonCrawl 2008 corpus already made significant gains over random initialization in both our benchmarks. This is despite using benchmark datasets from 2004 to 2012 which may differ stylistically from our pre-training corpus.

During model training, we used a train/dev split ratio of 0.9/0.1 and optimized the training loss with Adam \cite{Adam2014} on a learning rate of 0.001 and no dropout. We used a batch size of 48 webpage pairs, trained for 80 epochs and saved the best weights of the feature extractor to use in downstream benchmark tasks.

\subsection{Boilerplate Removal}

\begin{table}
    \caption{Boilerplate Removal}
    \label{tab:content extraction}
    \begin{tabular}{llcc}
        \toprule
        \multicolumn{2}{c}{\multirow{2}{*}{Model}} & \multicolumn{2}{c}{Micro-F1 (\%)}\\
        & & CleanEval & Dragnet\\
        \toprule
        \multicolumn{2}{l}{Perfect filter} & $97.9$ & $99.4$ \\
        \multicolumn{2}{l}{No filter} & $89.8$ & $64.9$ \\
        \midrule
        \multicolumn{2}{l}{python-readability \cite{python_readability2021}} & $90.2$ & $82.2$ \\
        \multicolumn{2}{l}{python-goose \cite{python_goose2015}} & $82.9$ & $76.1$ \\
        \multicolumn{2}{l}{CECTD-DS \cite{CETD}} & $92.3$\footnotemark & $82.8$ \\
        \multicolumn{2}{l}{Dragnet \cite{Dragnet2013}} & $91.2 \pm 0.1$ & $91.4 \pm 0.1$\\
        \multicolumn{2}{l}{Web2Text \cite{Web2Text2018}} & $93.0 \pm 0.1$ & $92.9 \pm 0.5$ \\
        \multicolumn{2}{l}{BoilerNet \cite{BoilerNet2020}} & $92.7 \pm 0.1$ & $93.8 \pm 0.3$ \\
        \midrule
        \multicolumn{2}{l}{$S=5, K=64$} & $92.8 \pm 0.2$ & $96.6 \pm 0.3$\\
        \multicolumn{2}{l}{$S=5, K=64$ + pre-train} & $93.0 \pm 0.3$ & $96.4 \pm 0.2$\\
        \multicolumn{2}{l}{$S=10, K=256$ w/o lstm} & $92.1 \pm 0.3$ & -\\
        \multicolumn{2}{l}{$S=10, K=256$ w/o residuals} & $91.3 \pm 0.7$ & -\\
        \multicolumn{2}{l}{$S=10, K=256$} & $92.7 \pm 0.2$ & $96.3 \pm 0.3$\\
        \multicolumn{2}{l}{$S=10, K=256$ + mask only} & $93.3 \pm 0.2$ & $96.8 \pm 0.2$\\
        \multicolumn{2}{l}{$S=10, K=256$ + pre-train} & $\mathbf{93.5 \pm 0.1}$ & $\mathbf{97.1 \pm 0.2}$\\
        \multicolumn{2}{l}{$S=5, T=5, K=256$} & $92.3 \pm 0.3$ & $95.6 \pm 0.4$\\
        \multicolumn{2}{l}{$S=5, T=5, K=256$ + mask only} & $93.2 \pm 0.2$ & $96.8 \pm 0.2$\\
        \midrule
        \multicolumn{2}{l}{$S=5, T=5, K=256$ + pre-train} & $93.4 \pm 0.2$ & $96.9 \pm 0.3$\\
        \bottomrule
    \end{tabular}
\end{table}

\footnotetext{This result differs substantially from the results in the original work. This could be due to a different test set used and some additional clean-up of the ground truth text that the authors performed.}

We benchmark our model against the previous state-of-the-art on the boilerplate removal task where the goal is to extract the main textual content from a webpage. We use 2 different datasets - CleanEval \cite{CleanEval2008} and Dragnet \cite{DragnetData2021}. The CleanEval corpus contains English webpages circa 2007 with a train/test split of 58/676 webpages. Dragnet is a larger corpus of webpages from 2012, and a train/test split of 966/415 webpages. Both datasets have accompanying extracted texts that represent the gold standard for each webpage.

While comparing the performance between different models on CleanEval, we note that some previous works are mutually incompatible in the use of metrics, train/test splits and even the objects of measurement. For example, CETD \cite{CETD} compares the extracted text based on the longest common subsequence (LCS) of word tokens with the ground truth, and because it relies on heuristics rather than machine learning, the authors evaluate on the entire CleanEval English corpus instead of just the test split. On the other hand, more recent works that utilize deep learning like Web2Text \cite{Web2Text2018} and BoilerNet \cite{BoilerNet2020} align the ground truth with individual text elements in the DOM to generate binary content vs boilerplate labels for each element, and then compare on the classification accuracy of these text elements based on the original test split. Other works report results using completely different splits with cross validation \cite{WebCEDOM2018}. %Yet others benchmark using a lexically unique bag-of-words comparison with the ground truth text \cite{Dragnet2013}\cite{CETR}, instead of the aforementioned LCS or other model specific metrics.

In order to compare apples-to-apples on a more consistent benchmark, we re-implement and re-evaluate past works on consistent dataset splits using appropriate metrics. We use the corpus’ original train/test splits for all the models and compare the extracted text with the ground truth text using the same LCS-based metrics described in the CETD work. Comparing the similarity of the extracted text directly with the corresponding ground truth text is less ambiguous than comparing accuracy on user-defined model outputs such as the predicted class of DOM text elements. After all, it is possible to have a high accuracy on model-specific classification outputs but ultimately still have a low similarity between the extracted text and the actual ground truth, and vice versa.

We describe the metric used in CETD \cite{CETD} that we adopt for this benchmark. Let $LCS(x, y)$ be the longest common subsequence of a pair of sequences $x, \, y$, then the metrics used for comparison between the sequences of (space delimited) word tokens of the extracted text $a$ and the ground truth text $b$ is:
\begin{align*}
    \text{Precision} = \dfrac{|LCS(a, b)|}{|a|}, \quad \text{Recall} = \dfrac{|LCS(a, b)|}{|b|}
\end{align*}

These metrics are first computed for each webpage, then averaged across all webpages and finally used  to derive the F1 score.

To use GROWN+UP for webpage boilerplate removal, we append a binary classifier at the output of the GROWN+UP feature extractor. Each node feature vector corresponding to each HTML tag containing text is passed through the classifier to predict whether a tag is content or boilerplate. Ground truth labels are generated for each HTML tag by aligning the raw ground truth text with the text within each HTML tag, similar to the methods used for other ML classifier models \cite{WebCEDOM2018,Dragnet2013}. Except for the removal of markup formatting tags in the CleanEval ground truth text, no other modifications to the raw HTML or ground truth texts are performed for either dataset.

For training on CleanEval, we initialize with the CommonCrawl 2008 pre-trained weights and fine-tune on the train set with a train/dev split of 53/5 webpages. We select the model based on the best F1 scores on the dev set after training for 40 epochs, which we then use to evaluate on the 676 test webpages. For the hyperparameters, we use a dropout of 0.3 and apply Adam with weight decay (AdamW) \cite{WeightDecay2019} with a learning rate of 0.002 and weight decay value of 0.0001. We also use label smoothing \cite{InceptionNet2016} with a value of 0.01 and a batch size of 128. We apply a similar methodology for Dragnet dataset but use a train/dev split of 869/97 webpages, a learning rate of 0.001 and no dropout.

In our benchmark, we repeat the entire experiment 10 times for stochastic models and compute statistics for each run\footnote{For simplicity, the assumption is made that the distribution of the sample means are approximately normal. This is quite presumptious since the sample size is small, but normality tests with larger samples seem to indicate some conformity in the underlying data at least for our models.}. For null hypothesis testing of the difference between two sample means across datasets or models, we first derive p-values using a two sample t-test in each case and then compute a global p-value using Fisher's combined probability test \cite{Fisher1954}.

The GROWN+UP variants compared in this benchmark are the $S = 5, T = 5, K = 256$ model that has 11.5M parameters, the $S = 10, K = 256$ model using only graph convolutions with roughly the same number of parameters, and the smaller $S = 5, K = 64$ with approximately 2.5M parameters. 

The results of this benchmark are shown in Table \ref{tab:content extraction}.

We first note that the maximum F1 score attainable even with a "perfect" boilerplate filter is less than 100\%. This is explained by sections of text in the ground truth that do not have an exact correspondence within the raw HTML, and therefore cannot be completely accounted for by just removing boilerplate elements within the DOM.

%Our pre-trained $S = 10, K = 256$ variant produces the highest mean F1 scores on both datasets, beating the previous state-of-the-art convincingly. This improvement is especially significant on the Dragnet dataset where most GROWN+UP model variants post noticeably better scores than the previous best even without pre-training.

From the benchmarks, our pre-trained $S = 5, T = 5, K = 256$ model posts noticeably better results over the previous state-of-the-art for both datasets, faring only slightly worse compared to our $S = 10, K = 256$ variant. This ablative comparison is within expectations since the Transformer blocks do not take graph edges into account during feature aggregation. Consequently, the neighborhood context around each graph node, likely crucial for node classification and for this particular benchmark, is mostly ignored.

Interestingly, there is evidence that our models pre-trained on both masking and same-website objectives perform better than the same model pre-trained only on the masking objective, given a global p-value of 0.005. This is a somewhat pleasant surprise since the same-website objective is hypothesized to improve downstream graph-level classification tasks rather than at the node-level. A simplistic explanation would be that low-level features learnt from the graph-level pre-training can be generally useful across tasks, but a more detailed study exploring transfer learning in this context is a subject for future work.

Also worth mentioning is the fact that pre-training on the smaller $S = 5, K = 64$ model seems to make little difference in the results based on a global p-value of 0.2 compared to random initialization, although the model still converges faster if pre-trained. This observation may be explained by the relatively smaller model capacity limiting the benefits of pre-training and may even have a negative impact as shown in past research \cite{BiT2020}.

Finally, the results lend support to our earlier claim that combining both residual layers and LSTMs in the $S = 10, K = 256$ model gives significantly better results than with either alone (combined p-value < 0.0001), at least on this benchmark.

\subsection{Genre Classification}

Unlike the webpage boilerplate removal task, the datasets used in previous works on webpage genre classification are extremely diverse \cite{WebPageClassSurvey2020}, with the vast majority proprietary or omitting the raw HTML sources \cite{DMOZ2016}. This makes a comprehensive benchmark on an appropriate standard somewhat challenging. In selecting a suitable dataset for experiments, we settled on the 7-Web-Genre (7-Web) \cite{Santini2008} and the KI-04 \cite{KI_042004} datasets for a few simple reasons: namely, that the datasets are one of the few that are still publicly available which include the full HTML sources and are also used by previous works in similar benchmarks. 7-Web is composed of 1400 webpages collected in 2005 and divided equally into 7 genres while KI-04 has 1209 webpages collected in 2004 divided into 8 genres.

For this benchmark, we attempt to classify the webpages in both datasets into their correct genres and measure model performance with 10-fold cross validation, following the common methodology in prior works \cite{Santini2008,KI_042004}.

To classify at the webpage or graph level, our model with $S = 5, T = 5$ produces a graph feature representation from the CLS node at the output of the last Transformer block. Alternatively, our $S = 10$ model appends a simple mean readout function at the output of the GROWN+UP feature extractor to pool features from all nodes to get a graph feature vector. We then apply Li-Arcface loss \cite{LiArcFace2019} to encourage better separation of class features. To reproduce results on our best model, we use AdamW and cosine decay with warm restarts \cite{SGDR2017} and apply an initial learning rate of 0.002 and decay steps at 5. For Li-Arcface loss parameters, we use a 5.0 scale factor and 0.3 margin. We also use 0.3 dropout to improve generalization performance and train for 35 epochs.

For our models, we repeat the 10-fold cross validation experiment 3 times with different folds per run to obtain statistics. We apply the paired difference t-test with corrected variance \cite{Nadeau2000} for null hypothesis testing of the cross validation performance between our models, and similar to the boilerplate removal benchmark, compute a global p-value using Fisher's method from the separate p-values.

\begin{table}
  \centering
  \caption{Genre Classification}
  \label{tab:genre}
  \begin{tabular}{llcc}
    \toprule
    \multicolumn{2}{c}{\multirow{2}{*}{Model}} & \multicolumn{2}{c}{Accuracy (\%)}\\
        & & 7-Web & KI-04\\
    \toprule
    \multicolumn{2}{l}{Santini \cite{Santini2008}} & $90.4$ & $68.9$\\
    \multicolumn{2}{l}{Meyer zu Eissen and Stein \cite{KI_042004}} & - & $70.0$\\
    \multicolumn{2}{l}{Kumari and Reddy \cite{PerfWebGenre2012}} & $91.5$ & -\\
    \multicolumn{2}{l}{Boese and Howe \cite{Boese2005}} & - & $74.8$\\
    \multicolumn{2}{l}{Kim and Ross \cite{Kim2007}} & $92.7$ & -\\
    \multicolumn{2}{l}{Mason et al. \cite{Mason2009}} & $94.6$ & -\\
    \multicolumn{2}{l}{Kanaris and Stamatatos \cite{Kanaris2009}} & 96.5 & 84.1\\
    \multicolumn{2}{l}{Deng et al. \cite{EnsembleWebGenre2020}} & 95.4 & - \\
    \midrule
    %GROWN+UP & \footnotesize{$S=5, K=64$} & $90.6 \pm 2.7$ & $68.6 \pm 5.1$\\
    %GROWN+UP & \footnotesize{$S=5, K=64$ + pt} & $91.0 \pm 2.2$ & $71.9 \pm 4.1$\\
    %\multicolumn{2}{l}{$S=5, K=64$ + pre-train} & $91.0 \pm 2.2$ & $71.9 \pm 4.1$\\
    \multicolumn{2}{l}{$S=10, K=256$ + mask only} & $93.8 \pm 1.8$ & $79.3 \pm 3.5$\\
    \multicolumn{2}{l}{$S=10, K=256$ + pre-train} & $94.8 \pm 1.7$ & $80.4 \pm 3.1$\\
    \multicolumn{2}{l}{$S=5, T=5, K=256$} & $85.9 \pm 3.1$ & $65.7 \pm 3.7$\\
    \multicolumn{2}{l}{$S=5, T=5, K=256$ + mask only} & $96.3 \pm 1.2$ & $84.5 \pm 2.1$\\
    \midrule
    \multicolumn{2}{l}{$S=5, T=5, K=256$ + pre-train} & $\mathbf{96.8 \pm 0.9}$ & $\mathbf{85.9 \pm 4.0}$\\
    \bottomrule
  \end{tabular}
\end{table}

We compare our general purpose model to other specialized genre classifiers, shown in Table \ref{tab:genre}.

Our pre-trained $S = 5, T = 5, K = 256$ model attains the best mean scores for both datasets, with the character n-gram model by Kanaris and Stamatatos \cite{Kanaris2009} in second place despite being a much simpler model than ours. It is however vital to note that unlike the n-gram based models \cite{Kanaris2009}\cite{Mason2009} which almost exclusively depend on textual features, our model also parses DOM structures. This capability may be especially relevant for distinguishing between webpages with little to no text, such as contemporary media-heavy websites.

We now examine some results from ablation experiments, in particular whether our claim that same-website pre-training improves downstream classification. Comparing the results on our model variants with pre-training on both objectives versus with masking only, there is good evidence (based on a global p-value of 0.041) that the same-website pre-training objective has a noticeable impact on this task, as hypothesized.

It is also apparent (global p-value of 0.0011) and expected that the addition of the Transformer block in the $S = 5, T = 5$ models gives substantial improvements over the simple mean readout used in the $S = 10$ models for both genre classification datasets.

Without pre-training, our $S = 5, T = 5, K = 256$ model converges to a very suboptimal result on both datasets, underlining the necessity of pre-training in this case. %Our smaller $S = 5, K = 64$ model however is able to converge to a baseline result without pre-training. In addition to converging faster with pre-training than without, there is also stronger evidence that pre-training on the smaller model is helpful on this genre classification task (based on global p-value of 0.057 at 10\% significance level), compared to the results in the content extraction benchmark.

\section{Conclusion and Future Work}
In our view, the research presented here paves the way towards a general parser of webpages and the web-at-large using deep learning. In addition to enabling webpage element level applications like web content scraping, we are also optimistic that a webpage can be distilled into a substantially smaller yet meaningful set of embeddings with GROWN+UP, similar to what has been done for images and text. This can facilitate interesting applications such as meta-analyses of a network of webpages and webpage indexing using representative embeddings.

%In our view, the research presented here paves the way towards a general parser of webpages and the web-at-large using deep learning. In the same manner that images and text documents can be reduced to substantially smaller yet meaningful set of application-agnostic feature embeddings, we are optimistic that webpages can also be distilled likewise with GROWN+UP, facilitating interesting applications such as meta-analyses of a network of websites, and webpage indexing and search using representative webpage embeddings. 

%We proposed a novel deep learning model inspired by the versatile and high capacity feature extractors popular in computer vision and NLP, that can parse webpages and can be applied to a variety of tasks in the space of web information retrieval. We also proposed a self-supervised pre-training algorithm that can scale to massive datasets inexpensively, and showed that a pre-trained model was able to attain respectable results in both web content extraction and web genre classification tasks using multiple datasets.

However, even though we have shown that our pre-trained model was able to attain promising results on two very different tasks using multiple datasets, a greater variety of experiments is required in order to establish its general applicability and performance across a wide spectrum of tasks. We release the source code of our model and benchmarking implementations for reference and to encourage further research in this area.

Moving forward, there are several clear avenues of exploration to further improve model performance generally. For instance, adding visual features from the DOM such as the bounding rectangle and computed styles of each element will likely yield significant improvements from the inclusion of important visual cues. Due to the significance of textual features in webpages, better sentence or paragraph representations \cite{Gao2021} may also have a large impact on overall performance. Lastly, integrating image, audio or video features in the model using the same approach as textual features will give a more holistic view of the webpage and will likely improve results, in addition to enabling more downstream applications like image content extraction.

\begin{acks}
This work was generously supported by Klass and we remain grateful for the free access to compute resources and the patience of our colleagues while we were hogging those. We also thank Lei Wang, Chen Kim Heng and other colleagues at Klass for their attentive reviews and useful feedback. 
\end{acks}

\bibliographystyle{ACM-Reference-Format}
\balance
\bibliography{main}

\end{document}